\documentclass[conference]{IEEEtran}
\IEEEoverridecommandlockouts
\usepackage{cite}
\usepackage{amsmath,amssymb,amsfonts}
\usepackage{algorithmic}
\usepackage{graphicx}
\graphicspath{ {./images/} }
\usepackage{textcomp}
\usepackage{xcolor}
\usepackage{subcaption} 
\def\BibTeX{{\rm B\kern-.05em{\sc i\kern-.025em b}\kern-.08em
    T\kern-.1667em\lower.7ex\hbox{E}\kern-.125emX}}
    
\usepackage[justification=centering]{caption}
\usepackage{float}

\usepackage{array}
\newcolumntype{M}[1]{>{\centering\arraybackslash}m{#1}}

\usepackage{enumitem}

\begin{document}

\title{TableNet: Deep Learning model for end-to-end Table detection and Tabular data extraction from Scanned Document Images\\
}

\author{
\IEEEauthorblockN{Shubham Paliwal, Vishwanath D, Rohit Rahul, Monika Sharma, Lovekesh Vig }
\IEEEauthorblockA{ TCS Research, New Delhi \\
\textit{\{shubham.p3, vishwanath.d2, monika.sharma1, rohit.rahul, lovekesh.vig\}}@tcs.com}
}

\maketitle

\begin{abstract}
With the widespread use of mobile phones and scanners to photograph and upload documents, the need for extracting the information trapped in unstructured document images such as retail receipts, insurance claim forms and financial invoices is becoming more acute.  A major hurdle to this objective is that these images often contain information in the form of tables and extracting data from tabular sub-images presents a unique set of challenges. This includes accurate detection of the tabular region within an image, and subsequently detecting and extracting information from the rows and columns of the detected table. While some progress has been made in table detection, extracting the table contents is still a challenge since this involves more fine grained table structure(rows \& columns) recognition. Prior approaches have attempted to solve the table detection and structure recognition problems independently using two separate models. In this paper, we propose TableNet: a novel end-to-end deep learning model for both table detection and structure recognition. The model exploits  the  interdependence between the twin tasks of table detection and table structure recognition to segment out the table and column regions. This is followed by semantic rule-based row extraction from the identified tabular sub-regions. The proposed model and extraction approach was evaluated on the publicly available ICDAR 2013 and Marmot Table datasets obtaining state of the art results. Additionally, we demonstrate that feeding additional semantic features further improves model performance and that the model exhibits transfer learning across datasets. Another contribution of this paper is to provide additional table structure annotations for the Marmot data, which currently only has annotations for table detection.

\end{abstract}


\section{Introduction}
With the proliferation of mobile devices equipped with cameras, an increasing number of customers are uploading documents via these devices, making the need for information extraction from these images more pressing. Currently, these document images are often manually processed resulting in high labour costs and inefficient data processing times. Frequently, these documents contain data stored in tables with multiple variations in layout and visual appearance. A key component of information extraction from these documents therefore involves digitizing the data present in these tabular sub-images. The variation in the table structure, and in the graphical elements  used to visually separate the tabular components make extraction from these images a very challenging problem. Most existing approaches to tabular information extraction  divide the problem into the two separate sub-problems of 1) table detection and 2) table structure recognition, and attempt to solve each sub-problem independently. While table detection involves detection of the image pixel coordinates containing the tabular sub-image, tabular structure recognition involves segmentation of the individual rows and columns in the detected table.  

In this paper, we propose TableNet, a novel end-to-end deep learning model that exploits the inherent interdependence between the twin tasks of table detection and table structure identification. The model utilizes a base network that is initialized with pre-trained VGG-19 features. This is followed by two decoder branches for 1) Segmentation of the table region and 2) Segmentation of the columns within a table region. Subsequently, rule based row extraction is employed to extract data in individual table cells. 

A multi-task approach is used for the training of the deep model. The model takes a single input image and produces two different semantically labelled output images for tables and columns. The model shares the encoding layer of VGG-19 for both the table and column detectors, while the decoders for the two tasks are separate. The shared common layers are repeatedly trained from the gradients received from both the table and column detectors while the decoders are trained independently. Semantic information about elementary data types is then utilized to further boost model performance. The utilization of the VGG-19 as a base network, which is pre-trained on the ImageNet dataset allows for exploitation of prior knowledge in the form of low level features learnt via training over ImageNet. 

We have evaluated TableNet's performance on the ICDAR-2013 dataset, demonstrating that our approach marginally outperforms other deep models as well as other state-of-the-art methods in detecting and extracting tabular information in image documents. We further demonstrate that the model can generalize to other datasets with minimal fine tuning, thereby enabling transfer learning. Furthermore, the Marmot dataset which has previously been annotated for table detection was also manually annotated for column detection, and these new annotations will be publicly released to the community for future research.

In summary, the primary contributions made in this paper are as follows:

\begin{enumerate}
\item We propose TableNet: a novel end-to-end deep multi-task architecture for both table detection and structure recognition yielding state of the art performance on the public benchmark ICDAR and Marmot datasets.
\item We demonstrate that  adding additional spatial semantic features to TableNet during training further boosts model performance.
\item We show that using a pre-trained TableNet model and fine tuning it on an another new dataset will boost the performance of the model on the new dataset, thereby allowing for transfer learning. 
\item We have manually annotated the Marmot dataset for table data extraction and will release the annotations to the community.
\end{enumerate}

The rest of the paper is organized as follows: Section \ref{sec:relwork} provides an overview of the related work on tabular information extraction. Section \ref{sec:TableNet} provides a detailed description of the TableNet model. Section \ref{sec:Extraction} outlines the extraction process with TableNet. 
Section \ref{sec:datasets} provides details about the datasets, preprocessing steps and training. Section \ref{sec:results} outlines the experiment details and results. Finally, the conclusions and future work are presented in Section \ref{sec:conc}. 
\section{Related Work}
\label{sec:relwork}
There is significant prior work on identifying and extracting the tabular data inside a document. Most of these have reported results on table detection and data extraction separately \cite{embley2006table}

Before the advent of deep learning, most of the work on table detection was based on heuristics or metadata. TINTIN \cite{pyreddy1997tintin} exploited structural information to identify tables and their component fields. \cite{cesarini2002trainable} used hierarchical representations based on the MXY tree for table detection and was the first attempt at using Machine Learning techniques for this problem.
T Kasar et al. \cite{kasar2013learning} identified intersecting horizontal, vertical lines and low-level features and used an SVM classifier to classify an image region as a table region or not.

Probabilistic graphical models were also used to detect tables; Silva et al. \cite{e2009learning} modelled the joint probability distribution over sequential
observations of visual page elements and the hidden state of a line (HMM) to merge potential table lines into tables resulted in a high degree of completeness. Jing Fang et al. \cite{fang2012table} used the table header as a starting point to detect the table region and decompose its elements. Raskovic et al. \cite{raskovic2018borderless} made an attempt to detect borderless tables. They utilized whitespaces as a heuristic rather than content for detection.

Recently, DeepDeSRT \cite{schreiber} was proposed which uses deep learning for both table detection and table structure recognition, i.e. identifying rows, columns, and cell positions in the detected tables. This work achieves state-of-the-art performance on the ICDAR 2013 table competition dataset. After this, \cite{kavasidis2018saliency} combined deep convolutional neural networks, graphical models and saliency concepts for localizing tables and charts in documents. This technique was applied on an extended version of ICDAR 2013 table competition dataset and outperforms existing models. \cite{tran2015table} locates the text components and extracts text blocks. After that, the height of each text block is compared with the average height and if satisfies a series of rules, the ROI is regarded as a table.

T-Recs \cite{kieninger1998t} was one of the earliest works to extract tabular data based on clustering of given word segments and overlap of the text inside the table. Y. Wang et al. \cite{wang2004table} estimates probabilities from geometric measurements made on the various entities in a given document.

Ashwin et al. \cite{tengli2004learning} exploit the formatting cues from semi-structured HTML tables to extract data from web pages. Here the cells are already demarcated by tags since they are in HTML tables. Singh et al. \cite{singh2018multidomain} use object detection techniques for Document Layout understanding.

\begin{figure*}[h]
\includegraphics[width=\textwidth]{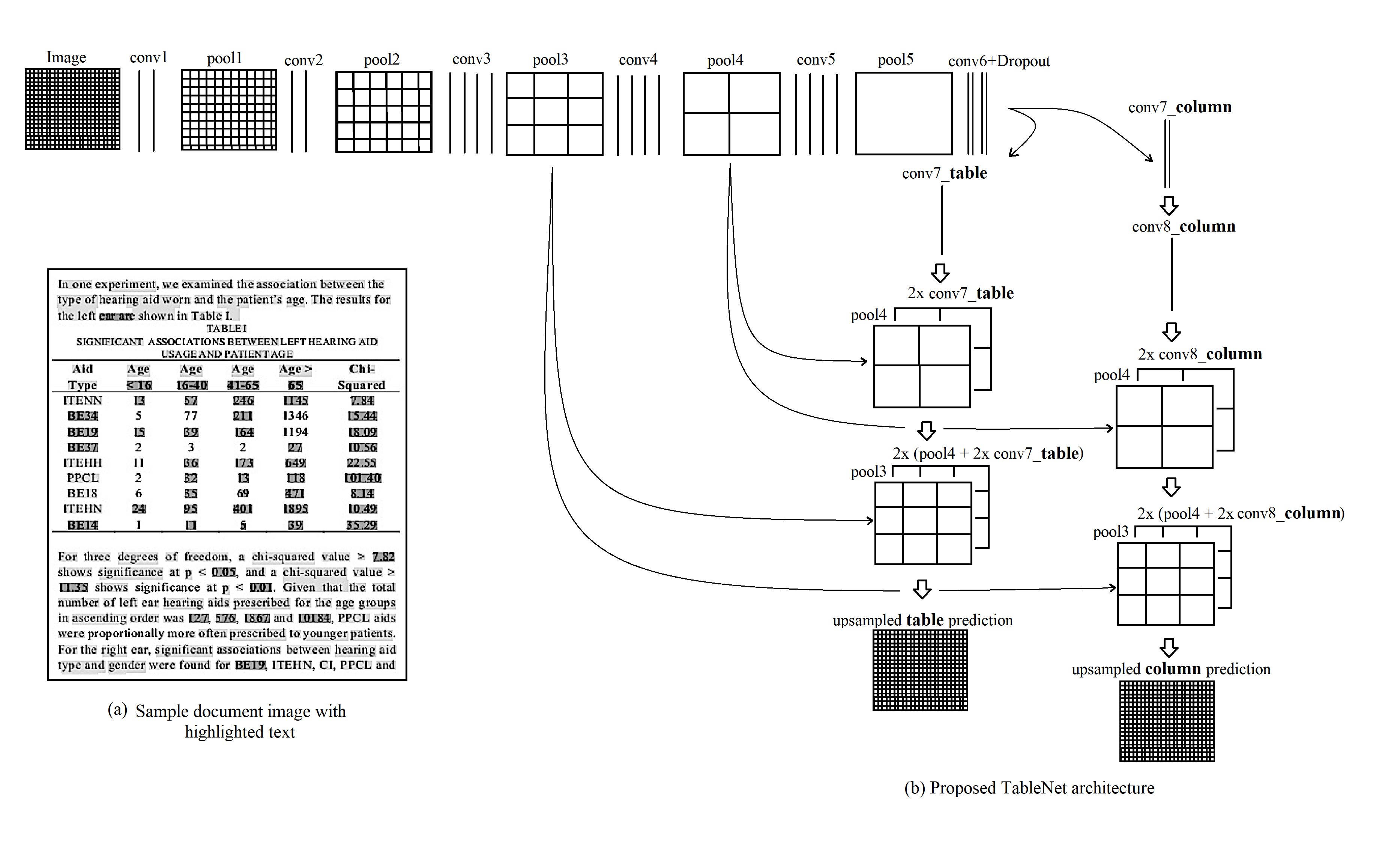}
\caption{(a) Sample training image from Marmot dataset, with highlighted text. (b) 
TableNet: Proposed model consists of pre-trained layers of VGG-19 as base network. Layers starting from conv1 to pool5 are used as common encoder layers for both table and column graph. Two decoder branches, conv7\_column and conv7\_table emerging after encoder layers, generate separate table predictions and column predictions.
}
\label{arch}
\end{figure*}

\section{TableNet: Deep Model for Table and Column Detection}
\label{sec:TableNet}
In all prior deep learning based approaches, table detection and column detection are considered separately as two different problems, which can be solved independently. However, intuitively if all the columns present in a document are known apriori, the table region can be determined easily. But by definition, columns are vertically aligned word/numerical blocks. Thus, independently searching for columns can produce a lot of false positives and knowledge of the tabular region can greatly improve results for column detection (since both tables and columns have common regions). Therefore, if convolutional filters utilized to detect tables, can be reinforced by column detecting filters, this should significantly improve the performance of the model. Our proposed model, exploits this intuition and  is based on the Long et al.\cite{DBLP:journals/corr/LongSD14}, encoder-decoder model for semantic segmentation. The encoder of the model is common across both tasks, but the decoder emerges as two different branches for tables and columns. Concretely, we enforced the encoding layers to use the ground truth of both tables and columns of document for training. However, the decoding layers are separated for table and column branches. Thus, there are two computational graphs to train.

The input image for the model, is first transformed into an RGB image and then, resized to 1024 * 1024 resolution. This modified image is processed using tesseract OCR \cite{smith2007overview} as described in the previous section. Since a single model produces both the output masks for the table and column regions, these two independent outputs have binary target pixel values, depending on whether the pixel region belongs to the table/column region or background respectively.

The problem of detecting  tables in documents is similar to the problem of detecting objects in real world images. Similar to the generic object detection problem, visual features of the tables can be used to detect tables and columns. The difference is that the tolerance for noise in table/column detection is much smaller than in object detection. Therefore, instead of regressing for the boundaries of tables and columns, we employed a  method to predict table and column regions pixel-wise. Recent work on semantic segmentation based on pixel wise prediction, has been very successful. FCN architecture, proposed by Long et al.\cite{DBLP:journals/corr/LongSD14}, has demonstrated the accuracy of encoder-decoder network architectures for semantic segmentation. The FCN architecture uses the skip-pooling technique to combine the low-resolution feature maps  of the decoder network with the high-resolution features of encoder networks. VGG-16 is used as the base layer in their model and fractionally-strided convolution layers are used to upscale the found low-resolution semantic map which is then combined with high resolution encoding layers.


\begin{figure*}
  \begin{subfigure}[b]{0.32\textwidth}
    \includegraphics[width=\textwidth]{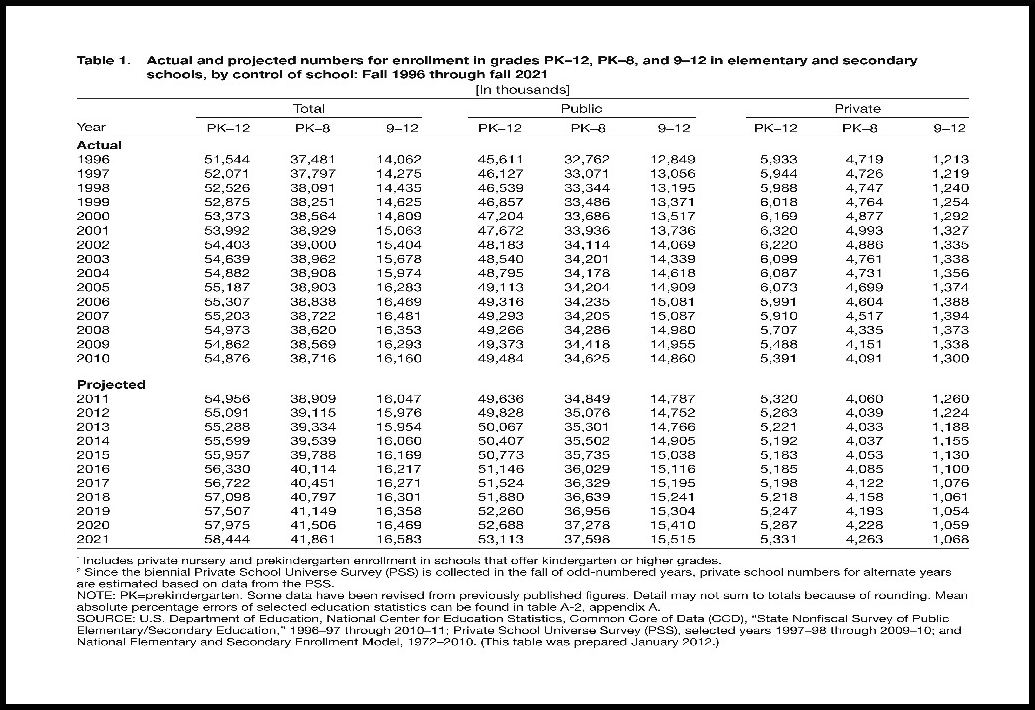}
    \caption{Figure showing the raw document image.}
    \label{fig:1}
  \end{subfigure}
  \begin{subfigure}[b]{0.32\textwidth}
    \includegraphics[width=\textwidth]{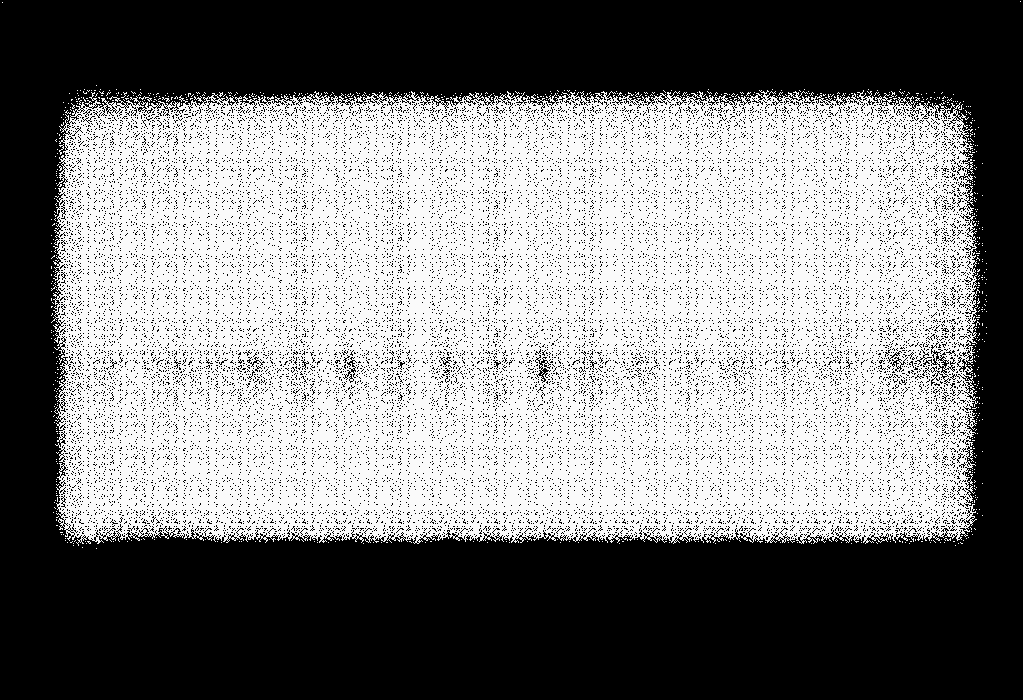}
    \caption{Generated table mask after processing.}
    \label{fig:2}
  \end{subfigure}
  \begin{subfigure}[b]{0.32\textwidth}
    \includegraphics[width=\textwidth]{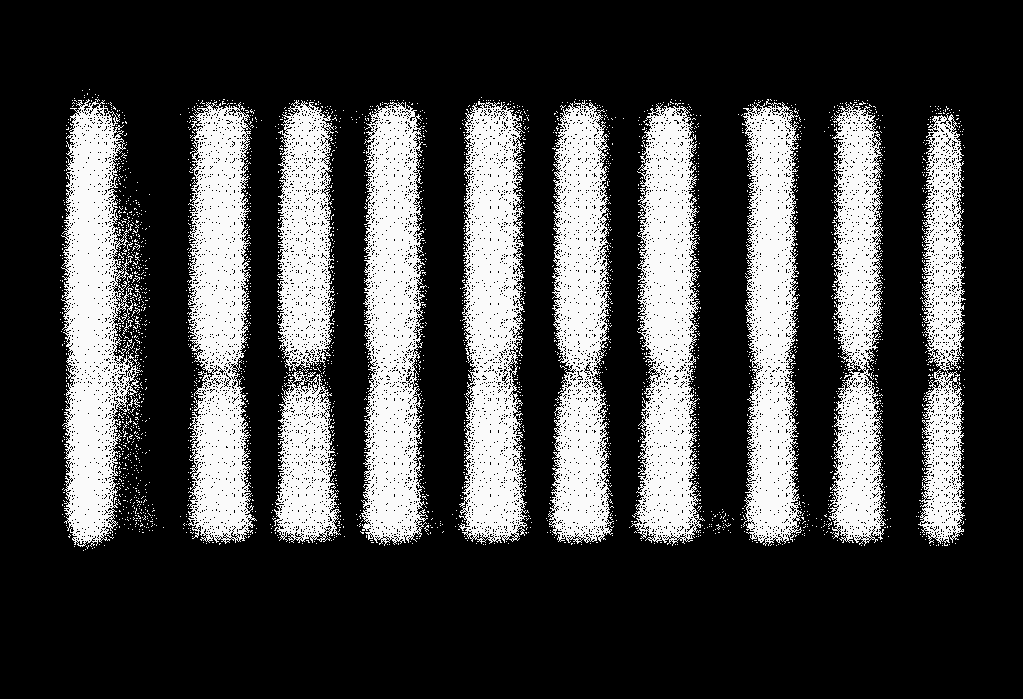}
    \caption{Generated column mask after processing.}
    \label{fig:3}
  \end{subfigure}
  \caption{Sample document image and its output masks generated after processing from TableNet.}
  \label{output}
\end{figure*}

Our model uses the same intuition for the encoder/decoder network as the FCN architecture. Our proposed model as shown in Figure \ref{arch}, uses a pre-trained VGG-19 layer as the base network. The fully connected layers (layers after pool5) of VGG-19 are replaced with two (1x1) convolution layers. Each of these convolution layers (conv6) uses the ReLU activation followed by a dropout layer having probability of 0.8 (conv6 + dropout as shown in Figure \ref{arch}). Following this layer, two different branches of the decoder network are appended. This is according to the intuition that the column region is a subset of the table region. Thus, the single encoding network can filter out the active regions with better accuracy using features of both table and column regions. The output of the (conv6 + dropout) layer is distributed to both decoder branches. In each branch, additional layers are appended to filter out the respective active regions. In the table branch of the decoder network, an additional (1x1) convolution layer, conv7\_table is used, before using a series of fractionally strided convolution layers for upscaling the image. The output of the conv7\_table layer is also up-scaled using fractionally strided convolutions, and is appended with the pool4 pooling layer of the same dimension. Similarly, the combined feature map is again up-scaled and the pool3 pooling is appended to it. Finally, the final feature map is upscaled to meet the original image dimension. In the other branch for detecting columns, there is an additional convolution layer (conv7\_column) with a ReLU activation function and a dropout layer with the same dropout probability. The feature maps are up-sampled using fractionally strided convolutions after a (1x1) convolution (conv8\_column) layer. The up-sampled feature maps are combined with the pool4 pooling layer and the combined feature map is up-sampled and combined with the pool3 pooling layer of the same dimension. After this layer, the feature map is up-scaled to the original image. In both branches, multiple (1x1) convolution layers are used before the transposed layers. The intuition behind using (1x1) convolution is to reduce the dimensions of feature maps (channels) which is used in class prediction of pixels, since the output layers (output of encoder network) must have channels equal to the number of classes (channel with max probability is assigned to corresponding pixels) which is later up-sampled. Therefore, the outputs of the two branches of computational graphs yield the mask for the table and column regions. 

\section{Table Row Extraction}
\label{sec:Extraction}
After processing the documents using TableNet, masks for table and column regions are generated. These masks are used to filter out the table and its column regions from the image. Since, all word positions of the document are already known (using Tesseract OCR), only the word patches lying inside table and column regions are filtered out. Now, using these filtered words, a row can be defined as the collection of words from multiple columns, which are at the similar horizontal level. However, a row is not necessarily confined to a single line, and depending upon the content of a column or line demarcations, a row can span multiple lines. Therefore, to cover the different possibilities, we formulate three rules for row segmentation:
\begin{itemize}
\item In most tables for which line demarcations are present, the lines segment the rows in each column. To detect the possible line demarcation (for rows), every space between two vertically placed words in a column is tested for the presence of lines via a Radon transform. The presence of horizontal line demarcation clearly segments out the row.
\item In case a row spans multiple lines, the rows of the table which have maximum non-blank entries is marked as the starting point for a new row. For example in a multi-column table, some of the columns can have entries spanning just one line (like quantity, etc), while others can have multi-line entries (like description, etc). Thus, each new row begins when all the entities in each column are filled.
\item In tables, where all the columns are completely filled and there are no line demarcations, each line (level) can be seen as a unique row.
\end{itemize}

\section{Dataset Preparation}
\label{sec:datasets}
Deep-learning based approaches are data-intensive and require large volumes of training data for learning effective representations. Unfortunately, there are very few datasets like Marmot\cite{6195411}, UW3\cite{Guyon97datasets}, etc for table detection and even these contain only a few hundred images. There are even fewer datasets for table structure identification such as the ICDAR 2013 table competition dataset for both table detection and its structural analysis \cite{gobel2013icdar}. This creates a constraint for deep learning models to solve both table detection and table structural analysis.

For training our model, we have used the Marmot table recognition dataset. This is the largest publicly available dataset for table detection but unfortunately did not have annotations for table columns or rows. We manually annotated the dataset for table structure recognition since the dataset had ground truth only for table detection. The dataset was manually annotated by labeling the bounding boxes around each of the columns inside the tabular region. The manually annotated modified dataset is publicly released with the name Marmot Extended for table structure recognition \footnote{\label{datasetlink}https://drive.google.com/drive/folders/1QZiv5RKe3xlOBdTzuTVuYRxixemVIOD}.

\subsection{Providing Semantic Information}
 Intuitively, any table has common data types in the same row/column depending on whether the table is in row major or column major form. For example, a name column will contain strings, while, a quantity header will contain numbers. To provide this semantic information to the deep model, text regions with similar data types are color coded. This modified image is then fed to the network resulting in improved performance.

We have included spatial semantic features by highlighting the words with patches as shown in Figure \ref{arch} (a) and this dataset is also made publicly available. The document images are first processed with histogram equalization. After pre-processing, the word blocks are extracted using tesseract OCR. These word patches are colored depending upon their basic data type. The resulting modified images are used as input to the network. The TableNet model takes the input image and generates the binary mask image of both table and columns separately. The achieved result is filtered using rules outlined on the basis of the detected table and column masks. An example of the generated output is shown in Figure \ref{output}.

\subsection{Training Data Preparation for TableNet}
\label{sec:TraningData}
To provide the basic semantic type information to the model, the word patches are color coded. The image is first processed with tesseract OCR, to get all the word patches in the image document. Then the words are processed via regular expressions to determine their data-type. The intuition is to color the word bounding boxes to impart both the semantic and spatial information to the network. Each data type is given a unique color and, bounding-boxes of words with similar data-types are shaded in the same color. Word bounding boxes are filtered out to remove spurious detections. However, since word detection and extraction from the OCR will produce some noise, the model needs to learn to be resilient to these cases. Therefore to simulate the case of incomplete word detection in the training image, a few randomly selected word patches are dropped deliberately. The formed color coded image can be used for training, but many visual features like line demarcations, corners, color highlights, etc are lost, while using only the word annotated document image. Therefore, to retain those important visual features in the training data, the word highlighted image is pixel-wise added to the original image. These modified document images are then used for training. 

\renewcommand{\arraystretch}{1.5}

\begin{table}
\centering

\begin{tabular}{|M{3.7cm}||M{1cm}|M{1cm}|M{1.3cm}|M{1.3cm}|  }
 \hline
Model &  Recall & Precision & F1-Score\\
\hline
\textbf{TableNet + Semantic Features (fine-tuned on ICDAR)}  &\textbf{0.9628} & \textbf{0.9697} & \textbf{0.9662}\\
\hline
\textbf{TableNet + Semantic Features} & \textbf{0.9621} & \textbf{0.9547} & \textbf{0.9583}\\
\hline
\textbf{TableNet} & \textbf{0.9501} & \textbf{0.9547} & \textbf{0.9547}\\
\hline
DeepDeSRT \cite{schreiber} & 0.9615 & 0.9740 & 0.9677 \\
\hline
Tran et al \cite{tran2015table} & 0.9636 & 0.9521 & 0.9578 \\
 \hline
\end{tabular}
\caption{Results on Table Detection} \label{tab:sometab1}
\end{table}





\begin{table}
\centering
\begin{tabular}{|M{3.7cm}||M{1cm}|M{1cm}|M{1.3cm}|M{1.3cm}|  }
 \hline
Model & Recall & Precision & F1-Score\\
\hline
\textbf{TableNet + Semantic Features (fine-tuned on ICDAR)} & \textbf{0.9001} & \textbf{0.9307} & \textbf{0.9151}\\
\hline
\textbf{TableNet + Semantic Features} & \textbf{0.8994} & \textbf{0.9255} & \textbf{0.9122}\\
\hline
\textbf{TableNet} & \textbf{0.8987} & \textbf{0.9215} & \textbf{0.9098}\\
\hline
DeepDeSRT \cite{schreiber} & 0.8736 & 0.9593 & 0.9144\\
\hline
\end{tabular}
\caption{Results on Table Structure Recognition \& Data Extraction} \label{tab:sometab2}
\end{table}

\section{EXPERIMENTS AND RESULTS}
\label{sec:results}
This section describes the different experiments performed on the ICDAR 2013 table competition dataset \cite{gobel2013icdar}. The model performance is evaluated based on the Recall, Precision \& F1-score. These measures are computed for each document and their average is taken across all the documents.

{

\begin{enumerate}[label=(\alph*)]
    \item Table Detection: Completeness and Purity are the two standard measures used in page segmentation \cite{e2011metrics}. A region is complete if it includes all sub-objects present in the ground-truth. A region is pure if it does not include any sub-objects which are not in the ground-truth. Sub-objects are created by dividing the given region into meaningful parts like heading of a table, body of a table etc. But these measures do not discriminate between minor and major errors. So, individual characters in each region are treated as sub-objects. Precision and recall measures are calculated on these sub-objects in each region and the average is taken across all the regions in a given document.
    \item Table Data Extraction: Each entry inside a table is defined as a cell. For each cell, adjacency relations are generated with the nearest horizontal and vertical neighbours. Thus, adjacency relations of a given cell is a 1D-tuple containing the text information of its neighboring cells. The content in each cell was normalized; white spaces were removed, special characters were replaced with underscores and lowercase letters with uppercase. This 1D-tuple can then be compared with the ground truth by using precision and recall.
\end{enumerate}

}   

TableNet requires both table and structure annotated data for training. We used the Marmot table detection data and manually annotated the structure information. There are a total of 1016 documents containing tables  including both Chinese and English documents, out of which 509 English documents are annotated and used for training. The proposed deep model has been implemented in Tensorflow and implemented on a system with  Intel(R) Xeon(R) Silver CPU having 32 cores and RAM of 128 GB Tesla V100-PCIE-1 GPU with 6GB of GPU memory. In Experiment 1, we trained our model on all positive samples of Marmot and tested on the ICDAR 2013 table competition dataset for both table and structure detection. There are two computation graphs which require training. Each training sample is a tuple of a document image, table mask and column mask. With each training tuple, the two graphs are computed at-least twice. In the initial phase of training, the table branch and column branch are computed in the ratio of 2:1. With each training tuple, the table branch of the computational graph is computed twice, and then the column branch of the model is computed. It is worth noting that, although the table branch and column branch are different, the encoder is the same for both. During initial iterations of training, the learning is more focused on training the model to generate big active tabular regions which on subsequent training specializes to column regions. After around 500 iterations with a batch size of 2, when train loss of both table and column detectors are comparable, this training scheme is modified. However, it should be noted that the table classifier at this stage must exhibit a converging trend (otherwise, training is extended with the same 2:1 scheme). The model is then trained in the ratio of 1:1 for both branches until convergence. Using the same training pattern, the model is trained for 5000 iterations with a batch size of 2 and learning rate of 0.0001. The Adam optimizer is used for improving and optimizing training with parameters beta1=0.9, beta2=0.999 and epsilon=1e-08. The convergence and over-fitting behavior was monitored by observing the performance over the validation set (taken from ICDAR 2013 data-set). During testing, 0.99 is taken as the threshold probability for pixel-wise prediction. The results are compiled in Table \ref{tab:sometab1} and Table \ref{tab:sometab2}. 

Similarly, in Experiment 2, we used the modified Marmot data-set where, the words in each document were highlighted to provide semantic context as described in Section \ref{sec:TraningData}. All the parameters were identical to Experiment 1. There was slight improvement in the results, when these spatial, semantic information are appended to the images  (see table for comparison). Output of the model is shown in Figure ~\ref{output}. Additionally, Experiment 3 was carried out to compare TableNet with the closest deep-learning based solution, DeepDSert \cite{schreiber}. In DeepDSert, separate models are made for Table detection and structure recognition, which were trained on different datasets such as Marmot for table detection, and the ICDAR 2013 table dataset for table structure recognition. To generate comparable results, we fine-tuned our Marmot trained TableNet model, on ICDAR train and test data splits. As done in DeepDSert, we also randomly chose 34 images for testing and used the rest of the data images for fine-tuning our model. Our model was fine-tuned, with the same parameters, in the ratio of 1:1 for both branches for 3000 iterations with a batch size of 2. The performance of our model further improved after the fine-tuning, possibly due to the introduction to the domain of ICDAR documents. The results of this experiments are also compiled in tables \ref{tab:sometab1} and \ref{tab:sometab2}. Average time taken for our system for each document image is 0.3765 seconds, however this could not be compared with DeepDSert as their model was not publicly available. While the results are not conclusively better than DeepDSert, they are certainly comparable and the fact that our model is end-to-end means further improvements can be made with richer semantic knowledge, and additional branches for learning row based segmentation.

\section{CONCLUSION}
\label{sec:conc}
This paper presents TableNet, a novel deep learning model trained on dual tasks of table detection and structure recognition in an end-to-end fashion. Existing approaches to information extraction treat detection and structure recognition  as two distinct problems to be solved independently. TableNet is the first model to jointly address both tasks simultaneously, by exploiting the inherent interdependence between table detection and table structure identification. TableNet utilizes the knowledge from previously learned tasks and can transfer that knowledge to newer, related ones demonstrating transfer learning. This is particularly useful when the training data is sparse. We also show that highlighting the text to provide data-type information improves the performance of the model. In the future we intend to train TableNet with a third branch to identify rows, however this will require a manual annotation exercise as currently datasets do not provide for row based annotations. Another question that arises from these experiments is what other semantic knowledge could be provided for better model performance, perhaps the use of more abstract data types such as currency, country or city,  might be useful.  

\bibliographystyle{IEEEtran}
\bibliography{file}

\end{document}